\documentclass[lettersize,journal]{IEEEtran}
\usepackage{amsmath,amsfonts}
\usepackage{algorithmic}
\usepackage{algorithm}
\usepackage{array}
\usepackage[caption=false,font=normalsize,labelfont=sf,textfont=sf]{subfig}
\usepackage{textcomp}
\usepackage{stfloats}
\usepackage{url}
\usepackage{verbatim}
\usepackage{graphicx}
\usepackage{cite}
\usepackage{arydshln}
\usepackage{xcolor}
\usepackage{bm}
\usepackage{multirow}
\usepackage{makecell}
\usepackage{booktabs}
\usepackage[hidelinks]{hyperref} 

\begin{document}

\title{GPA-RAM: Grasp-Pretraining Augmented Robotic Attention Mamba for Spatial Task Learning}

\author{Juyi Sheng$^{*}$, Yangjun Liu$^{*}$, Sheng Xu, \IEEEmembership{Senior Member, IEEE}, Zhixin Yang, \IEEEmembership{Senior Member, IEEE}, \\ Tiantian Xu, \IEEEmembership{Senior Member, IEEE}, Mengyuan Liu$^{\dagger}$, \IEEEmembership{Member, IEEE}  \\
\thanks{Juyi Sheng and Mengyuan Liu are with the State Key Laboratory of General Artificial Intelligence, Peking University, Shenzhen Graduate School, Shenzhen, 518055, China 
(email: liumengyuan@pku.edu.cn).
}
\thanks{Sheng Xu and Tiantian Xu are with the Shenzhen Institutes of Advanced Technology, Chinese Academy of Sciences, Shenzhen 518055, China 
}
\thanks{Yangjun Liu and Zhi-Xin Yang are with the State Key Laboratory of Internet of Things for Smart City, University of Macau, Macau 999078, China 
}
}

\markboth{IEEE/ASME Transactions on Mechatronics}%
{Shell \MakeLowercase{\textit{et al.}}: A Sample Article Using IEEEtran.cls for IEEE Journals}


\maketitle

\begin{abstract}
Fine-grained robotic manipulation often fails when inaccurate initial grasps propagate errors and necessitate complex pose correction. We propose Grasp-Pretraining Augmentation (GPA), which incorporates grasp priors from task demonstrations into imitation policies without additional grasp-pose data or annotation. When added to RVT2, GPA raises the average success rate on RLBench from 79.3\% to 84.2\%. When added to ACT, it raises success on ALOHA cube transfer and bimanual insertion from 86\% and 16\% to 98\% and 38\%, respectively. To offset added computational costs, we develop Robotic Attention Mamba (RAM) for real-time deployment. RAM combines attention-based spatial feature extraction with state-space modeling to capture long-range dependencies efficiently. The resulting GPA-RAM framework supports discrete keyframe prediction and continuous action generation. We evaluate it on four platforms, including physical UR5 and ARX R5 systems. GPA-RAM achieves an average success rate of 87.5\% on RLBench, outperforming RVT2 and ARP$^+$ by 8.2 and 2.6 percentage points, respectively. On ALOHA, it achieves 98\% success in cube transfer and 56\% in bimanual insertion, improvements of 12 and 40 percentage points over ACT, while operating at approximately 71 frames per second. These results demonstrate that GPA-RAM combines precise manipulation with efficient real-time robotic execution. Code is available at \url{https://gpa-ram.github.io/}.
\end{abstract}


\section{Introduction}

\IEEEPARstart{R}{obot} imitation learning (IL) enables manipulators to acquire visuomotor skills from demonstrations and reduces the need for task-specific programming \cite{tmech1}. Recent policies combine multi-view visual observations, language instructions, and robot proprioception to learn complex manipulation behaviors \cite{tmech2,tmech3,tmech4}. Transformer-based and diffusion-based approaches have improved the modeling of high-dimensional action distributions \cite{NIPS2017transformer,diffusion_model,tmech-dp}. Meanwhile, low-cost teleoperation systems such as ALOHA and UMI have reduced the barriers to collecting robot demonstrations \cite{aloha,umi}. Benchmarks such as RLBench and ALOHA further support the study of discrete keyframe prediction and continuous action generation \cite{james2020rlbench,aloha}. These advances have shifted the challenge from task completion alone to precise and responsive execution across robots and sensing configurations.

\begin{figure}[t]
    \centering
    \includegraphics[width=\linewidth]{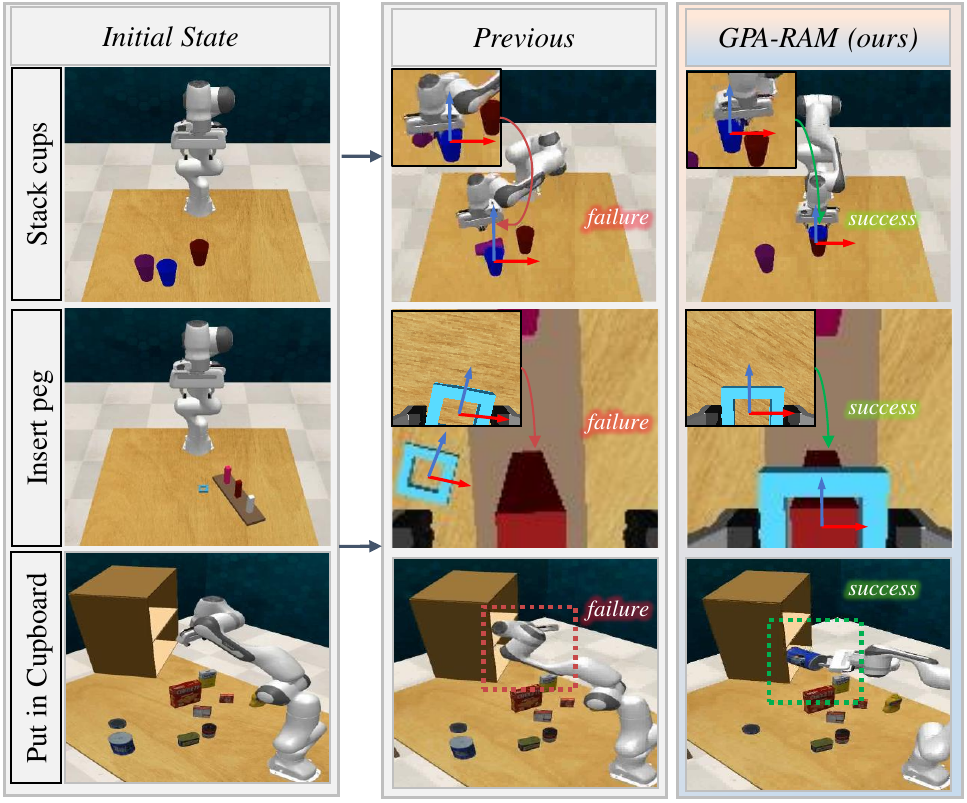}
    \caption{Representative manipulation failures of prior imitation-learning policies and the corresponding executions by GPA-RAM. Prior policies exhibit an early grasp failure during cup stacking (top), a grasp-induced pose error during peg insertion (middle), and a collision during cupboard placement (bottom). GPA-RAM completes the corresponding executions. These examples motivate the joint modeling of grasp-pose priors and scene-level spatial relations.}
    \label{fig:motivation}
\end{figure}

Despite this progress, task success remains sensitive to the initial grasp in operations requiring precise placement, alignment, or insertion. Methods such as PerACT, Act3D, PolarNet, RVT, RVT2, and ARP$^+$ learn task-level actions end to end \cite{shridhar2023perceiver,gervet2023act3d,chen23polarnet,goyal2023rvt,goyal2024rvt2,arp}. However, their policy representations do not explicitly preserve grasp-pose priors for downstream manipulation. An inaccurate grasp can terminate a task, as in cup stacking, or produce a pose error that prevents subsequent insertion (Fig.~\ref{fig:motivation}). Correcting such errors may require regrasping or in-hand adjustment, which is difficult with limited demonstrations and parallel-jaw grippers. The resulting challenge is how to strengthen grasp-sensitive representations without collecting and annotating a separate grasp dataset.

To address this challenge, we propose Grasp-Pretraining Augmentation (GPA). GPA pretrains a grasp-pose detector using grasp configurations already contained in expert demonstrations. The detector output head is subsequently removed, and the retained grasp features are fused with the task-policy features during IL. This design introduces an explicit manipulation prior without additional data collection or manual grasp-pose annotation. Because GPA can be integrated with both RVT2 \cite{goyal2024rvt2} and ACT \cite{aloha}, it serves as a modular augmentation for discrete and continuous manipulation policies.

Improved grasp perception alone, however, does not resolve the computational requirements of real-time robotic execution. The additional grasp-perception branch increases latency, while multi-view RGB-D observations produce long spatial token sequences. Many existing RLBench methods rely on Transformer blocks to fuse visual, linguistic, and proprioceptive information \cite{shridhar2023perceiver,goyal2023rvt,goyal2024rvt2,arp}. The computational cost of global self-attention grows quadratically with sequence length \cite{transformer_shortcoming}. In contrast, state-space models and Mamba provide linear-time sequence modeling for long input sequences \cite{gu2022efficientlymodelinglongsequences,gu2023mamba}. We therefore develop Robotic Attention Mamba (RAM), which retains attention for view-specific feature interaction and employs Mamba blocks to model long-range spatial dependencies. This hybrid design reduces dependence on global self-attention while preserving the spatial reasoning required for precise action prediction. A coarse-to-fine RAM predicts discrete keyframe actions, whereas a single-stage variant supports continuous action generation. Together, GPA and RAM balance grasp precision, spatial reasoning, and inference efficiency.

The resulting GPA-RAM framework is evaluated on four robot platforms with different sensing and actuation configurations. The experiments cover Franka Panda and ALOHA systems in simulation, together with physical UR5 and ARX R5 robots. This evaluation examines both discrete and continuous manipulation under simulated and physical operating conditions. The main contributions are summarized as follows:
\begin{itemize}
    \item We propose Grasp-Pretraining Augmentation (GPA), a modular framework that incorporates grasp-pose priors into robot learning networks to improve precise grasping without additional data collection or pose annotation.

    \item We develop Robotic Attention Mamba (RAM), a hybrid architecture that combines attention mechanisms with linear-time state-space modeling for efficient spatial feature extraction and real-time action generation.

    \item We validate GPA-RAM across diverse tasks on four simulated and physical robot systems. GPA-RAM achieves an average success rate of 87.5\% on RLBench, exceeding RVT2 and ARP$^+$ by 8.2 and 2.6 percentage points, respectively. On ALOHA, it improves cube-transfer and bimanual-insertion success over ACT by 12 and 40 percentage points while running at approximately 71 FPS.
\end{itemize}

\section{Related Work}

\subsection{Vision-Based Robotic Manipulation}

Vision-based robotic manipulation has progressed from grasp detection to end-to-end policy learning. Early approaches focused on detecting stable grasp poses from visual observations \cite{grasp1,grasp2,grasp3,tmech5}. Multi-stage pipelines decompose manipulation into perception, planning, and execution \cite{yang2023watch_and_act,zhao2023twostageinsert,tro1}. This decomposition supports complex tasks, while errors at early stages can propagate through the pipeline.

End-to-end methods such as ALOHA and Diffusion Policy directly map visual observations to robot actions \cite{aloha,chi2023diffusionpolicy}, while UMI reduces demonstration collection costs \cite{umi}. These policies are commonly optimized for individual tasks, increasing data collection and retraining costs as the task set expands. Recent approaches use video data \cite{video1,video2,video3,video4,v1} or target images \cite{img2img1,img2img2} to improve skill transfer across scenarios. Their task-level objectives leave initial grasp quality implicit, limiting precision in grasp-sensitive placement and insertion tasks.

\subsection{Transformer-Based Robot Learning}

Transformer-based policies provide a unified mechanism for integrating visual, linguistic, and proprioceptive information. PerACT encodes language and voxelized scene representations for 6-DOF manipulation but incurs substantial training and computation costs \cite{shridhar2023perceiver}. Language-conditioned policies such as BC-Z, HULC, and $\Sigma$-agent improve task adaptation \cite{lang1,lang2,lang3}. RVT and RVT2 reduce 3D processing costs through multi-view rendering \cite{goyal2023rvt,goyal2024rvt2}, whereas ARP$^+$ predicts action components autoregressively \cite{arp}. RT-1, RT-2, and FP2AT further extend Transformer policies to large-scale or multimodal robot control \cite{RT1,RT2,fp2at}.

The computational cost of global self-attention grows quadratically with token sequence length, increasing latency for high-resolution multi-view observations \cite{transformer_shortcoming}. Moreover, task-level action learning keeps grasp-pose priors implicit, making high-precision manipulation sensitive to the initial grasp. GPA-RAM combines explicit grasp-prior augmentation with attention-Mamba spatial modeling to improve manipulation precision and inference efficiency.
\begin{figure*}[ht]
    \centering
    \includegraphics[width=0.9\linewidth,keepaspectratio]{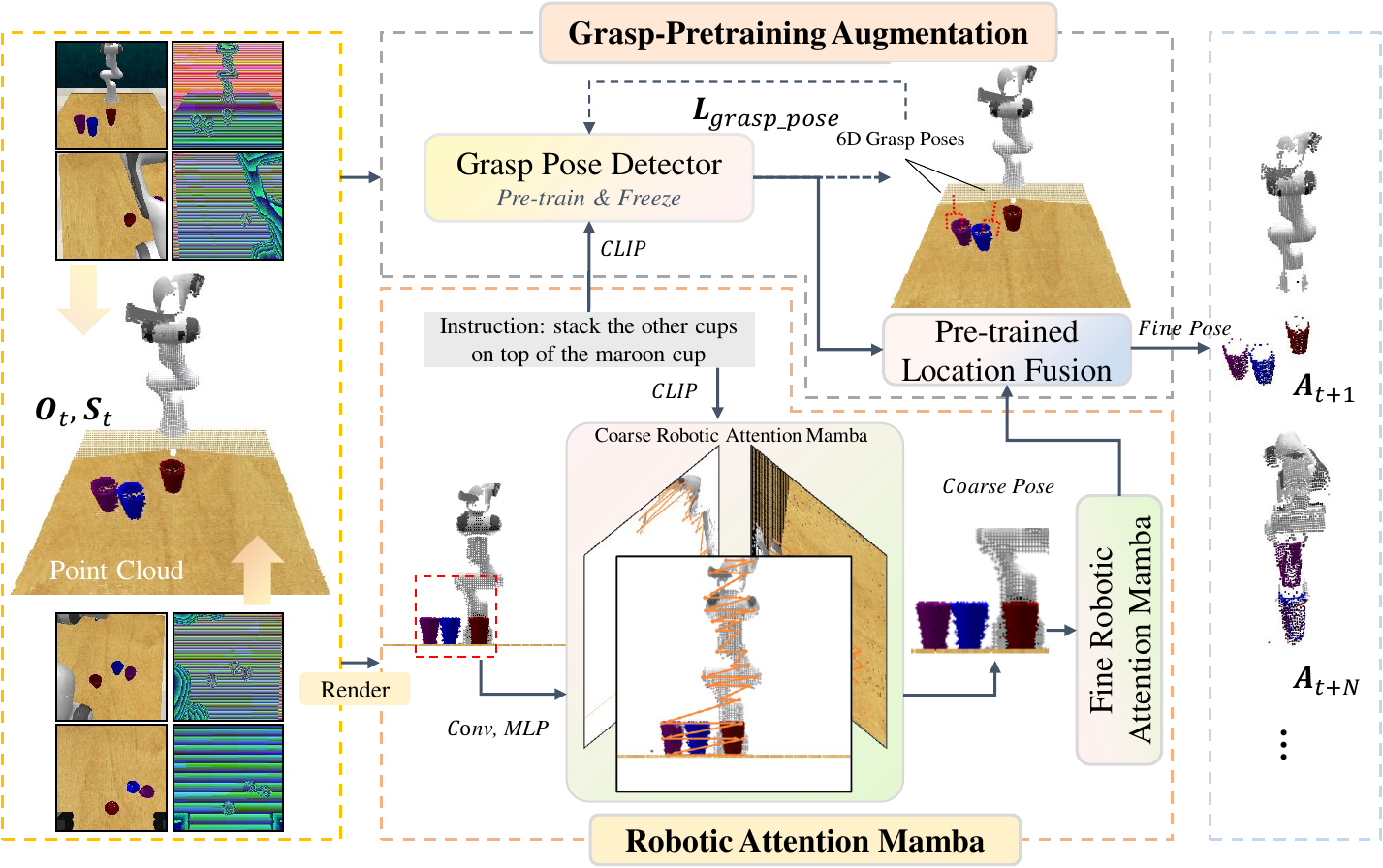}  
    \caption{Overview of the GPA-RAM on RLBench. The GPA-RAM primarily comprises two modules: RAM for spatial feature extraction and GPA for grasping skills. It processes point clouds, RGB-D, language instructions, and robot proprioception, extracting spatial semantic features through Coarse and Fine Robotic Attention Mamba in RAM and grasping features via the pre-trained \& frozen Grasp Pose Detector in GPA. These features are fused in GPA's Pre-trained Location Fusion to predict the next key action in 3D space.
    }
    \label{fig:overview}
\end{figure*}

\section{Proposed GPA-RAM Framework}

Figure~\ref{fig:overview} presents the overall architecture of GPA-RAM. The framework comprises GPA for grasp-prior augmentation and RAM for efficient spatial feature extraction. GPA transfers grasp-pose features learned from task demonstrations to the manipulation policy. RAM models spatial relations among multi-view observations, language instructions, and robot states. Two policy variants are constructed for different action-generation settings. The discrete policy employs a coarse-to-fine RAM architecture to predict keyframe actions, whereas the continuous policy uses a single-stage RAM architecture to generate action sequences.

\subsection{Preliminaries}

\textbf{Problem Formulation and Keyframe Extraction.}
In RLBench \cite{james2020rlbench}, a multi-task policy is trained to solve $M$ manipulation tasks, including closing a jar, inserting an object onto a square peg, and opening a drawer. For task $k$, the demonstration dataset is defined as
\begin{equation}
    \mathcal{D}_k=\left\{\mathbf{d}_i^k\right\}_{i=1}^{N},
\end{equation}
where $N$ denotes the number of expert demonstrations. Each demonstration contains a sequence of camera transformations, language instructions, visual observations, robot actions, and proprioceptive states:
\begin{equation}
    \mathbf{d}_i^k=
    \left\{
    \left(
    \mathbf{T}_t^i,
    \mathbf{L}_k,
    \mathbf{O}_t^i,
    \mathbf{A}_t^i,
    \mathbf{S}_t^i
    \right)
    \right\}_{t=1}^{n_i},
\end{equation}
where $n_i$ is the original sequence length. The camera transformations are represented by
\begin{equation}
    \mathbf{T}_t^i\in
    \mathbb{R}^{n_{\mathrm{cams}}\times4\times4},
\end{equation}
where $n_{\mathrm{cams}}$ is the number of cameras. These transformations map the camera observations to a common 3D coordinate frame.

The language instruction $\mathbf{L}_k$ specifies the current task, such as ``open the top drawer.'' The visual observation
\begin{equation}
    \mathbf{O}_t^i\in
    \mathbb{R}^{n_{\mathrm{cams}}\times7\times H\times W}
\end{equation}
contains RGB, depth, and XYZ point-coordinate channels from each camera, where $H$ and $W$ denote the image height and width. The robot action is represented as
\begin{equation}
    \mathbf{A}_t^i=
    \left[
    \mathbf{a}_{\mathrm{trans},t}^i,
    \mathbf{a}_{\mathrm{rot},t}^i,
    a_{\mathrm{gripper},t}^i,
    a_{\mathrm{collision},t}^i
    \right],
\end{equation}
where $\mathbf{a}_{\mathrm{trans},t}^i\in\mathbb{R}^{3}$ and $\mathbf{a}_{\mathrm{rot},t}^i\in\mathbb{R}^{3}$ denote the end-effector translation and rotation. The variables $a_{\mathrm{gripper},t}^i$ and $a_{\mathrm{collision},t}^i$ represent the binary gripper and collision-control states. The proprioceptive state $\mathbf{S}_t^i$ contains the current robot-state information.

Following previous RLBench methods \cite{shridhar2023perceiver,gervet2023act3d,goyal2024rvt2,fp2at}, frames with near-zero robot velocity or a change in gripper state are selected as keyframes. The resulting keyframe sequence is
\begin{equation}
    \widetilde{\mathbf{d}}_i^k=
    \left\{
    \left(
    \mathbf{T}_j^i,
    \mathbf{L}_k,
    \mathbf{O}_j^i,
    \mathbf{A}_j^i,
    \mathbf{S}_j^i
    \right)
    \right\}_{j=1}^{m_i},
\end{equation}
where $m_i\leq n_i$. At keyframe $j$, the policy predicts the next keyframe action from the current observation, proprioceptive state, and language instruction:
\begin{equation}
    \mathbf{A}_{j+1}
    =
    \Phi_{\theta}
    \left(
    \mathbf{O}_j,
    \mathbf{S}_j,
    \mathbf{L}_k
    \right),
\end{equation}
where $\Phi_{\theta}$ denotes the multi-task policy parameterized by $\theta$.

\textbf{Action Discretization.}
Each training sample contains the current observation tuple
$\left(\mathbf{O}_j,\mathbf{S}_j,\mathbf{L}_k\right)$
and the next keyframe action $\mathbf{A}_{j+1}$. The translation target remains a 3D position. The three Euler-angle components are independently quantized into 72 bins, corresponding to a resolution of $5^{\circ}$ per bin. The discretized rotation label is represented as
\begin{equation}
    \overline{\mathbf{a}}_{\mathrm{rot}}
    =
    \left[
    \mathbf{v}^{x},
    \mathbf{v}^{y},
    \mathbf{v}^{z}
    \right]
    \in\{0,1\}^{3\times72},
\end{equation}
where $\mathbf{v}^{q}$ is the one-hot label for rotation axis $q\in\{x,y,z\}$. For each axis,
\begin{equation}
    \sum_{f=1}^{72}v_f^{q}=1,
    \qquad
    v_f^{q}\in\{0,1\}.
\end{equation}
The gripper and collision-control targets are binary:
\begin{equation}
    \hat{a}_{\mathrm{gripper}},
    \hat{a}_{\mathrm{collision}}
    \in\{0,1\}.
\end{equation}
The complete ground-truth action is therefore written as
\begin{equation}
    \mathbf{A}_{j+1}
    =
    \left[
    \hat{\mathbf{a}}_{\mathrm{trans}},
    \overline{\mathbf{a}}_{\mathrm{rot}},
    \hat{a}_{\mathrm{gripper}},
    \hat{a}_{\mathrm{collision}}
    \right].
\end{equation}

\textbf{State-Space Modeling.}
State-space models represent an input sequence through a latent state. A continuous-time linear state-space model is written as
\begin{align}
    \dot{\mathbf{h}}(t)
    &=
    \mathbf{A}\mathbf{h}(t)
    +
    \mathbf{B}\mathbf{x}(t),\\
    \mathbf{y}(t)
    &=
    \mathbf{C}\mathbf{h}(t),
\end{align}
where $\mathbf{x}(t)\in\mathbb{R}^{D}$, $\mathbf{h}(t)\in\mathbb{R}^{N}$, and $\mathbf{y}(t)\in\mathbb{R}^{D}$ denote the input, latent state, and output, respectively. The system matrices satisfy
\begin{equation}
    \mathbf{A}\in\mathbb{R}^{N\times N},
    \quad
    \mathbf{B}\in\mathbb{R}^{N\times D},
    \quad
    \mathbf{C}\in\mathbb{R}^{D\times N}.
\end{equation}

Using zero-order hold with a discretization step $\Delta$, the continuous system is converted into
\begin{align}
    \overline{\mathbf{A}}
    &=
    \exp\left(\Delta\mathbf{A}\right),\\
    \overline{\mathbf{B}}
    &=
    \mathbf{A}^{-1}
    \left[
    \exp\left(\Delta\mathbf{A}\right)-\mathbf{I}
    \right]
    \mathbf{B},\\
    \mathbf{h}_t
    &=
    \overline{\mathbf{A}}\mathbf{h}_{t-1}
    +
    \overline{\mathbf{B}}\mathbf{x}_t,\\
    \mathbf{y}_t
    &=
    \mathbf{C}\mathbf{h}_t.
\end{align}

Mamba introduces input-dependent state-space parameters to control how information is retained across a sequence \cite{gu2023mamba}. For a batch size $B_s$, sequence length $L$, and feature dimension $D$, the discretization parameter satisfies
\begin{equation}
    \Delta\in\mathbb{R}^{B_s\times L\times D}.
\end{equation}
The parameters $\Delta_t$, $\mathbf{B}_t$, and $\mathbf{C}_t$ are generated from the input token, while $\mathbf{A}$ remains a shared learnable parameter. The selective state update is expressed as
\begin{align}
    \mathbf{h}_t
    &=
    \overline{\mathbf{A}}_t\mathbf{h}_{t-1}
    +
    \overline{\mathbf{B}}_t\mathbf{x}_t,\\
    \mathbf{y}_t
    &=
    \mathbf{C}_t\mathbf{h}_t.
\end{align}
This input-dependent parameterization selectively retains task-relevant information and supports linear-time sequence processing.

\subsection{Overall Network Architecture}

GPA-RAM comprises two complementary modules: RAM for efficient spatial feature extraction and GPA for incorporating grasp-pose priors. Figure~\ref{fig:overview} presents the GPA-RAM architecture for discrete keyframe prediction on RLBench, and Fig.~\ref{fig:network} details the internal structures of RAM and GPA. For discrete tasks, GPA-RAM employs coarse-to-fine spatial prediction. For continuous tasks, a single-stage RAM and GPA are integrated with an action-chunking decoder.

\subsection{Robotic Attention Mamba}
\label{sub-RAM}

RAM combines attention-based feature interaction with state-space sequence modeling. As shown in Fig.~\ref{fig:overview}, multi-view RGB-D observations are first converted into a point cloud $\mathbf{P}_t$. The point cloud is then rendered from $V$ predefined virtual viewpoints:
\begin{equation}
    \mathbf{O}_t^v
    =
    \mathcal{R}
    \left(
    \mathbf{P}_t,\mathbf{C}_v
    \right),
    \qquad v\in\{1,\ldots,V\},
\end{equation}
where $\mathcal{R}$ denotes the rendering operation and $\mathbf{C}_v$ represents the parameters of virtual camera $v$.

RAM adopts a coarse-to-fine architecture for discrete keyframe prediction. In the coarse stage, an MLP encodes the robot proprioceptive state, a convolutional encoder extracts virtual-view features, and CLIP \cite{radford2021clip} encodes the language instruction:
\begin{equation}
\begin{aligned}
    \mathbf{F}_s
    &=
    \operatorname{MLP}_s(\mathbf{S}_t),\\
    \mathbf{F}_O^v
    &=
    \operatorname{Conv}
    \left(
    \mathbf{O}_t^v
    \right),\\
    \mathbf{F}_L
    &=
    \operatorname{CLIP}(\mathbf{L}_k).
\end{aligned}
\end{equation}

The coarse RAM module, denoted by $\phi_{\mathrm{C\text{-}RAM}}$, processes the visual, language, and proprioceptive features. A prediction head generates coarse heatmaps for the virtual views:
\begin{equation}
    \mathbf{Q}_C
    =
    \operatorname{Upsample}
    \left[
    \operatorname{MLP}_C
    \left(
    \phi_{\mathrm{C\text{-}RAM}
    }
    \left(
    \{\mathbf{F}_O^v\}_{v=1}^{V},
    \mathbf{F}_s,
    \mathbf{F}_L
    \right)
    \right)
    \right].
\end{equation}

The coarse heatmaps are decoded using the virtual-camera geometry to obtain a coarse 3D position:
\begin{equation}
    \widehat{\mathbf{p}}_C
    =
    \operatorname{Decode3D}
    \left(
    \mathbf{Q}_C,
    \{\mathbf{C}_v\}_{v=1}^{V}
    \right).
\end{equation}
A local point-cloud region centered at $\widehat{\mathbf{p}}_C$ is extracted and re-rendered to produce fine-scale observations $\mathbf{O}_t^{\prime v}$. The fine RAM module then generates the final spatial representation:
\begin{equation}
\begin{aligned}
    \mathbf{F}_O^{\prime v}
    &=
    \operatorname{Conv}
    \left(
    \mathbf{O}_t^{\prime v}
    \right),\\
    \mathbf{F}_{\mathrm{RAM}}
    &=
    \phi_{\mathrm{F\text{-}RAM}}
    \left(
    \{\mathbf{F}_O^{\prime v}\}_{v=1}^{V},
    \mathbf{F}_s,
    \mathbf{F}_L
    \right).
\end{aligned}
\label{eq:f_ram}
\end{equation}

\begin{figure}[t]
    \centering
    \includegraphics[width=\linewidth]{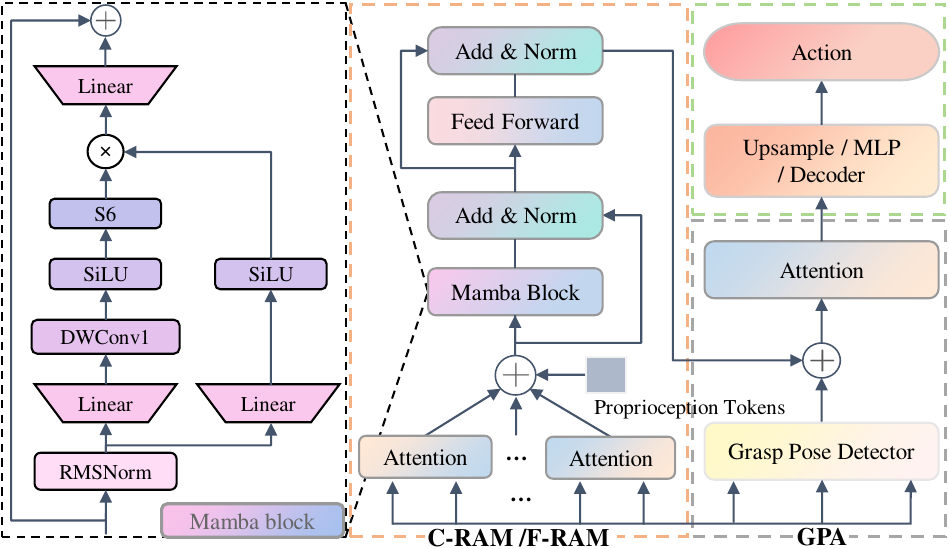}
    \caption{Architectures of RAM and GPA. Left: internal structure of a Mamba block. Middle: a RAM module processes visual patches and robot proprioceptive tokens using per-view attention and state-space sequence modeling. Right: GPA extracts grasp features using a pretrained grasp-pose detector and combines them with RAM features through PLoFusion for action prediction.}
    \label{fig:network}
\end{figure}

As illustrated in Fig.~\ref{fig:network}, each RAM stage first applies self-attention within each virtual view to model interactions among visual patches. The resulting view features are combined with the language and proprioceptive features. Mamba blocks process the flattened tokens as one-dimensional sequences and capture long-range spatial dependencies. The computational cost of each Mamba block scales linearly with token-sequence length. Layer normalization, feedforward layers, and residual connections are applied to produce the spatial representation.

\subsection{Grasp-Pretraining Augmentation}

GPA incorporates grasp-pose priors extracted from expert demonstrations into the manipulation policy. Before policy training, a grasp-pose detector $\psi_G$ is trained using grasp poses contained in the task demonstrations. This process uses the existing demonstrations without additional data collection or manual grasp-pose annotation.

The pretrained detector extracts grasp-sensitive features from the current visual observation and language instruction:
\begin{equation}
    \mathbf{F}_G
    =
    \psi_G
    \left(
    \mathbf{O}_t,
    \mathbf{F}_L
    \right).
\end{equation}
During policy training, the detector output head is removed, and its pretrained feature backbone is frozen. The Pre-trained Location Fusion module, denoted by $\phi_{\mathrm{PLF}}$, spatially aligns the grasp and RAM features before fusion:
\begin{equation}
    \mathbf{F}_{\mathrm{GPA}}
    =
    \phi_{\mathrm{PLF}}
    \left(
    \mathbf{F}_G,
    \mathbf{F}_{\mathrm{RAM}}
    \right).
\label{eq:gpa_fusion}
\end{equation}

PLoFusion concatenates the aligned features and applies self-attention to model interactions between grasp-sensitive local features and scene-level spatial features. The resulting representation $\mathbf{F}_{\mathrm{GPA}}$ is used by the action-prediction heads.

\subsection{Action Prediction}
\label{sub-AP}

Separate prediction heads map $\mathbf{F}_{\mathrm{GPA}}$ to the translation heatmap, rotation logits, gripper state, and collision-control state:
\begin{equation}
\begin{aligned}
    \mathbf{Q}_F
    &=
    \operatorname{Upsample}
    \left[
    \operatorname{MLP}_{\mathrm{trans}}
    \left(
    \mathbf{F}_{\mathrm{GPA}}
    \right)
    \right],\\
    \mathbf{Z}_{\mathrm{rot}}
    &=
    \operatorname{MLP}_{\mathrm{rot}}
    \left(
    \mathbf{F}_{\mathrm{GPA}}
    \right),\\
    z_{\mathrm{gripper}}
    &=
    \operatorname{MLP}_{\mathrm{gripper}}
    \left(
    \mathbf{F}_{\mathrm{GPA}}
    \right),\\
    z_{\mathrm{collision}}
    &=
    \operatorname{MLP}_{\mathrm{collision}}
    \left(
    \mathbf{F}_{\mathrm{GPA}}
    \right).
\end{aligned}
\end{equation}

At inference time, the fine heatmaps are decoded into a 3D translation:
\begin{equation}
    \mathbf{a}_{\mathrm{trans}}^{\mathrm{pred}}
    =
    \operatorname{Decode3D}
    \left(
    \mathbf{Q}_F
    \right).
\end{equation}
The three rotation components are obtained from the maximum-logit categories. The binary gripper and collision-control states are obtained by thresholding their sigmoid probabilities. The predicted keyframe action is
\begin{equation}
    \mathbf{A}_{t+1}^{\mathrm{pred}}
    =
    \left[
    \mathbf{a}_{\mathrm{trans}}^{\mathrm{pred}},
    \mathbf{a}_{\mathrm{rot}}^{\mathrm{pred}},
    a_{\mathrm{gripper}}^{\mathrm{pred}},
    a_{\mathrm{collision}}^{\mathrm{pred}}
    \right].
\end{equation}

The translation loss is computed between the predicted heatmaps and the ground-truth heatmaps:
\begin{equation}
    \mathcal{L}_{\mathrm{trans}}
    =
    \operatorname{CE}
    \left(
    \mathbf{Q}_F,
    \widehat{\mathbf{Q}}_F
    \right),
\end{equation}
where $\widehat{\mathbf{Q}}_F$ is generated by projecting the ground-truth translation onto the virtual views. The rotation loss is the sum of the classification losses for the three Euler-angle components:
\begin{equation}
    \mathcal{L}_{\mathrm{rot}}
    =
    \sum_{q\in\{x,y,z\}}
    \operatorname{CE}
    \left(
    \mathbf{Z}_{\mathrm{rot}}^q,
    \widehat{\mathbf{v}}^q
    \right).
\end{equation}

The gripper and collision-control losses are binary cross-entropies:
\begin{align}
    \mathcal{L}_{\mathrm{gripper}}
    &=
    \operatorname{BCE}
    \left(
    \sigma(z_{\mathrm{gripper}}),
    \hat{a}_{\mathrm{gripper}}
    \right),\\
    \mathcal{L}_{\mathrm{collision}}
    &=
    \operatorname{BCE}
    \left(
    \sigma(z_{\mathrm{collision}}),
    \hat{a}_{\mathrm{collision}}
    \right).
\end{align}

The total objective is
\begin{equation}
\begin{aligned}
    \mathcal{L}_{\mathrm{total}}
    ={}&
    \lambda_{\mathrm{trans}}
    \mathcal{L}_{\mathrm{trans}}
    +
    \lambda_{\mathrm{rot}}
    \mathcal{L}_{\mathrm{rot}}\\
    &+
    \lambda_{\mathrm{gripper}}
    \mathcal{L}_{\mathrm{gripper}}
    +
    \lambda_{\mathrm{collision}}
    \mathcal{L}_{\mathrm{collision}},
\end{aligned}
\end{equation}
where the $\lambda$ terms control the contribution of each prediction objective. The model parameters are optimized as
\begin{equation}
    \theta^*
    =
    \arg\min_{\theta}
    \mathcal{L}_{\mathrm{total}}.
\end{equation}

\subsection{Adaptation to Continuous-Action Tasks}

The ALOHA bimanual benchmark requires continuous action generation at $50\,\mathrm{Hz}$. Each demonstration contains a sequence of visual observations, robot actions, and proprioceptive states:
\begin{equation}
    \mathbf{d}_i
    =
    \left\{
    \left(
    \mathbf{O}_t,
    \mathbf{A}_t,
    \mathbf{S}_t
    \right)
    \right\}_{t=1}^{n_i}.
\end{equation}

At each time step, the policy predicts an action chunk of length $H_a$:
\begin{equation}
    \mathbf{A}_{t:t+H_a-1}^{\mathrm{pred}}
    =
    \widetilde{\Phi}_{\theta}
    \left(
    \mathbf{O}_t,
    \mathbf{S}_t
    \right).
\end{equation}

\begin{figure}[t]
    \centering
    \includegraphics[width=\linewidth]{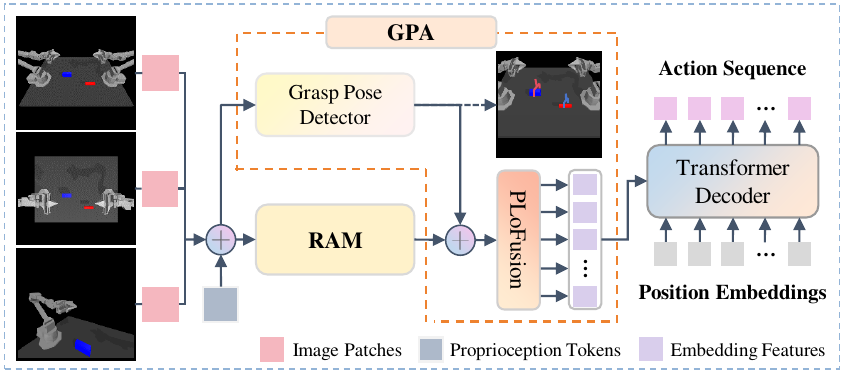}
    \caption{Adaptation of GPA and single-stage RAM to continuous action generation. Visual and proprioceptive features are encoded by RAM and fused with grasp-sensitive features from GPA. A Transformer decoder subsequently generates continuous action chunks.}
    \label{fig:act_new}
\end{figure}

As shown in Fig.~\ref{fig:act_new}, GPA and RAM are integrated with the ACT architecture through two modifications. First, the Transformer encoder is replaced by a single-stage RAM. The coarse-to-fine rendering procedure is omitted because continuous control requires a feature representation at every time step:
\begin{equation}
\begin{aligned}
    \widetilde{\mathbf{F}}_O
    &=
    \operatorname{Conv}
    \left(
    \mathbf{O}_t
    \right),\\
    \widetilde{\mathbf{F}}_{\mathrm{RAM}}
    &=
    \widetilde{\phi}_{\mathrm{RAM}}
    \left(
    \widetilde{\mathbf{F}}_O,
    \operatorname{MLP}(\mathbf{S}_t)
    \right).
\end{aligned}
\end{equation}

Second, a ResNet-based grasp-pose encoder extracts grasp-sensitive visual features:
\begin{equation}
    \widetilde{\mathbf{F}}_G
    =
    \widetilde{\psi}_G
    \left(
    \mathbf{O}_t
    \right).
\end{equation}
PLoFusion combines the grasp and RAM features:
\begin{equation}
    \widetilde{\mathbf{F}}_{\mathrm{GPA}}
    =
    \widetilde{\phi}_{\mathrm{PLF}}
    \left(
    \widetilde{\mathbf{F}}_G,
    \widetilde{\mathbf{F}}_{\mathrm{RAM}}
    \right).
\end{equation}

A Transformer decoder uses learned action-query embeddings $\mathbf{E}$ to generate the future action chunk:
\begin{equation}
    \mathbf{A}_{t:t+H_a-1}^{\mathrm{pred}}
    =
    \operatorname{TransformerDecoder}
    \left(
    \widetilde{\mathbf{F}}_{\mathrm{GPA}},
    \mathbf{E}
    \right).
\end{equation}

The continuous-action loss is averaged across the predicted action chunk:
\begin{equation}
    \mathcal{L}_{\mathrm{ALOHA}}
    =
    \frac{1}{H_a}
    \sum_{h=0}^{H_a-1}
    \left\|
    \mathbf{A}_{t+h}^{\mathrm{pred}}
    -
    \mathbf{A}_{t+h}
    \right\|_2^2.
\end{equation}
The continuous policy is trained by
\begin{equation}
    \widetilde{\theta}^{*}
    =
    \arg\min_{\widetilde{\theta}}
    \mathcal{L}_{\mathrm{ALOHA}}.
\end{equation}

\section{Experimental Evaluation}

We evaluated GPA-RAM in CoppeliaSim, MuJoCo, and physical environments. The evaluation involved four robot platforms: simulated Franka Emika Panda and ALOHA dual-arm systems, together with physical UR5 and ARX R5 systems. The experiments covered discrete keyframe-based manipulation on RLBench and continuous bimanual manipulation in both simulated and physical settings. The evaluation considered task success, inference speed, and performance across different robot and camera configurations.

\begin{table*}[ht]
\caption{Multi-task performance evaluation on RLBench. The table summarizes the success rates across 18 manipulation tasks. GPA-RAM achieves a SOTA average success rate of 87.5\%, outperforming previous baselines such as RVT2 and ARP$^+$. 
}
\vspace{-4mm}
\label{table:rlbench}
\begin{center}
\resizebox{\textwidth}{!}{
\begin{tabular}{lcccccccccccccccccc} 
\toprule
Model  & \makecell{Avg. \\ Success $\uparrow$} & \makecell{Avg. \\ Rank $\downarrow$} & \makecell{Close \\ Jar} & \makecell{Drag \\ Stick} & \makecell{Insert \\ Peg} & \makecell{Meat off \\Grill} & \makecell{Open \\ Drawer} & \makecell{Place \\Cups} & \makecell{Place \\ Wine} & \makecell{Push \\ Buttons} \\
\midrule
Image-BC (CNN)~\cite{lang1} & 1.3 & 12.14 & 0 & 0 & 0 & 0 & 4 & 0 & 0 & 0 \\
Image-BC (ViT)~\cite{lang1} & 1.3 & 12.31 & 0 & 0 & 0 & 0 & 0 & 0 & 0 & 0 \\
C2F-ARM-BC~\cite{james2022coarse} & 20.1 & 11.08 & 24 & 24 & 4 & 20 & 20 & 0 & 8 & 72 \\
HiveFormer~\cite{guhur2023hiveformer} & 45.3 & 9.25 & 52.0 & 76.0 & 0.0 & \textbf{100.0} & 52.0 & 0.0 & 80 & 84 \\
PerACT~\cite{shridhar2023perceiver} & 49.4 & 7.94 & 55.2 $\pm$ 4.7 & 89.6 $\pm$ 4.1 & 5.6 $\pm$ 4.1 & 70.4 $\pm$ 2.0 & 88.0 $\pm$ 5.7 & 2.4 $\pm$ 3.2 & 44.8 $\pm$ 7.8 & 92.8 $\pm$ 3.0 \\
PolarNet~\cite{chen23polarnet} & 46.4 & 8.33 & 36.0 & 92.0 & 4.0 & \textbf{100.0} & 84.0 & 0.0 & 40 & 96 \\
Act3D~\cite{gervet2023act3d} & 65.0 & 5.83 & 92.0 & 92.0 & 27.0 & 94.0 & \textbf{93.0} & 3.0 & 80 & 99 \\
RVT~\cite{goyal2023rvt} & 62.9 & 6.61 & 52.0 $\pm$ 2.5 & 99.2 $\pm$ 1.6 & 11.2 $\pm$ 3.0 & 88.0 $\pm$ 2.5 & 71.2 $\pm$ 6.9 & 4.0 $\pm$ 2.5 & 91.0 $\pm$ 5.2 & 92.0 $\pm$ 0.0 \\
RVT2~\cite{goyal2024rvt2} & 79.3 & 4.53 & 99.0 $\pm$ 1.0 & 98.0 $\pm$ 2.0 & 43.0 $\pm$ 7.7 & 98.0 $\pm$ 2.0 & 78.0 $\pm$ 8.3 & 28.0 $\pm$ 7.5 & 95.0 $\pm$ 1.7 & 96.0 $\pm$ 0.0 \\

ARP \cite{arp} & 81.6 & 4.19 & 97.6 &88.0 &53.2 &96.0 &90.4 &48 &92.0 &\textbf{100.0} \\
ARP$^+$ \cite{arp}            & 84.9                    & 3.28                      & 95.2                      & 99.2                      & 78.4                      & 97.6                    & 92.8                    & \textbf{48.8}            & 96.0                      & \textbf{100.0}              \\
RAM (ours) & 84.7 & \textbf{3.22} & \textbf{100.0} $\pm$ 0.0 & \textbf{100.0} $\pm$ 0.0 & 77.0 $\pm$ 1.7 & 98.0 $\pm$ 2.0 & 78.0 $\pm$ 6.6 & 47.0 $\pm$ 3.3 & 93.0 $\pm$ 3.3 & 96.0 $\pm$ 0.0 \\
GPA-RAM (ours) & \textbf{87.5} & \textbf{2.28} & \textbf{100.0 }$\pm$ 0.0 & \textbf{100.0} $\pm$ 0.0  & \textbf{95.0} $\pm$ 1.7 & \textbf{100.0} $\pm$ 0.0 & 84.0 $\pm$ 6.7 & 40.0 $\pm$ 4.0 & \textbf{100.0} $\pm$ 0.0 & 89.8 $\pm$ 3.2 \\
\midrule
Model & \makecell{Put in \\ Cupboard} & \makecell{Put in \\ Drawer} & \makecell{Put in \\ Safe} & \makecell{Screw \\ Bulb} & \makecell{Slide \\ Block} & \makecell{Sort \\ Shape} & \makecell{Stack \\ Blocks} & \makecell{Stack \\ Cups} & \makecell{Sweep to \\ Dustpan} & \makecell{Turn \\ Tap} \\
\midrule
Image-BC (CNN)~\cite{lang1} & 0 & 8 & 4 & 0 & 0 & 0 & 0 & 0 & 0 & 8 \\
Image-BC (ViT)~\cite{lang1} & 0 & 0 & 0 & 0 & 0 & 0 & 0 & 0 & 0 & 16 \\
C2F-ARM-BC~\cite{james2022coarse} & 0 & 4 & 12 & 8 & 16 & 8 & 0 & 0 & 0 & 68 \\
HiveFormer~\cite{guhur2023hiveformer} & 32.0 & 68.0 & 76.0 & 8.0 & 64.0 & 8.0 & 8.0 & 0.0 & 28.0 & 80 \\
PerACT~\cite{shridhar2023perceiver} & 28.0 $\pm$ 4.4 & 51.2 $\pm$ 4.7 & 84.0 $\pm$ 3.6 & 17.6 $\pm$ 2.0 & 74.0 $\pm$ 13.0 & 16.8 $\pm$ 4.7 & 26.4 $\pm$ 3.2 & 2.4 $\pm$ 2.0 & 52.0 $\pm$ 0.0 & 88.0 $\pm$ 4.4 \\
PolarNet~\cite{chen23polarnet} & 12.0 & 32.0 & 84.0 & 44.0 & 56.0 & 12.0 & 4.0 & 8.0 & 52.0 & 80 \\
Act3D~\cite{gervet2023act3d} & 51.0 & 90.0 & 95.0 & 47.0 & 93.0 & 8.0 & 12.0 & 9.0 & 92.0 & 94 \\
RVT~\cite{goyal2023rvt} & 49.6 $\pm$ 3.2 & 88.0 $\pm$ 5.7 & 91.2 $\pm$ 3.0 & 48.0 $\pm$ 5.7 & 81.6 $\pm$ 5.4 & 36.0 $\pm$ 2.5 & 28.8 $\pm$ 3.9 & 26.4 $\pm$ 8.2 & 72.0 $\pm$ 0.0 & 93.6 $\pm$ 4.1 \\
RVT2~\cite{goyal2024rvt2} & 59.0 $\pm$ 4.4 & 96.0 $\pm$ 0.0 & 96.0 $\pm$ 4.0 & 89.0 $\pm$ 3.3 & 81.0 $\pm$ 3.3 & 47.0 $\pm$ 3.3 & 62.0 $\pm$ 6.0 & 77.0 $\pm$ 3.3 & 98.0 $\pm$ 2.0 & 87.5 $\pm$ 3.7 \\
ARP \cite{arp} & 68.0 & 99.2 &94.4 &85.6 & \textbf{98.4} & 35.2 & 55.2 &76.8 &90.4 &\textbf{100.0} \\
ARP$^+$ \cite{arp} & 69.6 & 98.4& 86.4  & 89.6  & 92.8                      & 46.4 & 63.2                      & 80.0                      & 97.6                      & 96.0                \\
RAM (ours) & \textbf{79.0} $\pm$ 4.4 & \textbf{100.0} $\pm$ 0.0 & 96.0 $\pm$ 2.8 & 88.0 $\pm$ 0.0 & 91.0 $\pm$ 1.7 & 55.0 $\pm$ 5.2 & \textbf{70.4} $\pm$ 5.4 & 62.0 $\pm$ 6.6 & \textbf{100.0} $\pm$ 0.0 & 94.0 $\pm$ 3.5 \\
GPA-RAM (ours) & 70.0 $\pm$ 2.0 & \textbf{100.0} $\pm$ 0.0 & \textbf{98.0} $\pm$ 2.0 & \textbf{92.0} $\pm$ 0.0 & 93.0 $\pm$ 1.7 & \textbf{66.0} $\pm$ 2.0 & 65.6 $\pm$ 2.0 & \textbf{84.8} $\pm$ 1.6 & \textbf{100.0} $\pm$ 0.0 & 97.0 $\pm$ 3.3  \\
\bottomrule
\end{tabular}
}
\end{center}
\vspace{-4mm}
\end{table*}

\subsection{Benchmarks in Simulation} 
In simulation, this paper evaluated the proposed method using the RLBench \cite{james2020rlbench} and ALOHA \cite{aloha} benchmarks. The RLBench benchmark, simulated on CoppeliaSim platform, includes 18 tasks, where existing methods frequently encounter difficulties in high-precision grasping and placement tasks, such as inserting a ring onto a square peg, and placing a triangular prism into a shape sorter. Each demonstration consists of RGB-D images from the front, left shoulder, right shoulder, and wrist camera views, a language description, and the states of the Franka Panda robot at each time step. Keyframe discovery, selecting critical task frames (e.g., zero velocity or gripper state change), is exploited to reduce data redundancy and improve model learning efficiency following \cite{shridhar2023perceiver,gervet2023act3d,goyal2023rvt,goyal2024rvt2}. The ALOHA benchmark, generated on MuJoCo \cite{mujoco} platform, contains two fine-grained bimanual tasks: a cube transfer task and an insertion task. Each demonstration provides RGB images from multiview cameras and the states of the ALOHA dual-arm robot at each time step.

\subsection{Baselines} For RLBench, the proposed GPA-RAM was compared with leading methods, including those tailored directly for 3D visual observations such as 3D point clouds and voxel grids (C2F-ARM-BC \cite{james2022coarse, shridhar2023perceiver}, PerACT \cite{shridhar2023perceiver}, PolarNet \cite{chen23polarnet}, and Act3D \cite{gervet2023act3d}), and the latest image-based approaches (Image-BC (CNN) \cite{lang1, shridhar2023perceiver}, Image-BC (ViT) \cite{lang1, shridhar2023perceiver}, HiveFormer \cite{guhur2023hiveformer}, RVT \cite{goyal2023rvt}, RVT2 \cite{goyal2024rvt2}) and ARP$^+$ \cite{arp}. They are all based on Transformer or CNN. The proposed method is a Mamba-based method introduced in this multi-task benchmark. For ALOHA dataset, we adapted the proposed RAM and GPA into Action Chunking with Transformers \cite{aloha} (abbreviated as ACT, and different from Act3D), and compared them against the top-performing methods, i.e., BC-ConvMLP \cite{lang1}, BeT \cite{bet}, RT-1 \cite{RT1}, VINN \cite{VINN}, and ACT \cite{aloha}. 

\subsection{Implementation Details} 
The proposed GPA-RAM and the most competitive baseline are trained and tested on three RTX 4090 GPUs for the RLBench tasks.
The training is conducted with a batch size of 24 over 99 epochs, starting with a learning rate of 0.0018, which is dynamically adjusted using the Cosine Annealing Learning Rate scheduler. Each task is trained on 100 demonstrations and tested four times on 25 unseen scenarios, using models from the final 30 training epochs. The best-performing model is selected for the final success rate comparison.
For the ALOHA tasks, training is performed on a single RTX 4090 GPU. Each task is trained with 50 scripted demonstrations, and the best model during training is chosen for testing. Each task is tested with 50 different trials. Since the ALOHA dataset lacks language features, the language components in GPA-RAM are replaced with the features of the current robot arm state.
In our implementation, we employ M2T2 \cite{yuan2023m2t2} as the grasp pose detector for RLBench tasks, which are trained in a multi-task manner. Conversely, for ALOHA tasks, we use ResNet \cite{he2016deep} and a single-task training approach. We primarily report Inference Speed on the continuous ALOHA benchmark, where real-time control is especially crucial.

\subsection{Evaluation Metrics}

To rigorously evaluate our models against prior work, we report performance across 18 tasks from the RLBench benchmark. We employ two key metrics to measure performance:

\begin{itemize}
    \item \textbf{Average Success Rate (\%):} This is the primary metric for overall performance. It is calculated as the mean percentage of successful episodes across 18 tasks. A higher value indicates better overall task-solving capability.
    
    \item \textbf{Average Rank (Avg. Rank):} To assess the consistency and comprehensive performance of each model across the full task suite, we also compute the Average Rank. For each task, all models are ranked based on their success rate (e.g., 1st, 2nd, 3rd...). We then average these ranks across all 18 tasks for each model. A \textbf{lower} Average Rank signifies a more robust and consistently high-performing model across the entire benchmark.
\end{itemize}
\begin{figure*}[htbp]
    \centering
    \includegraphics[width=1\linewidth]{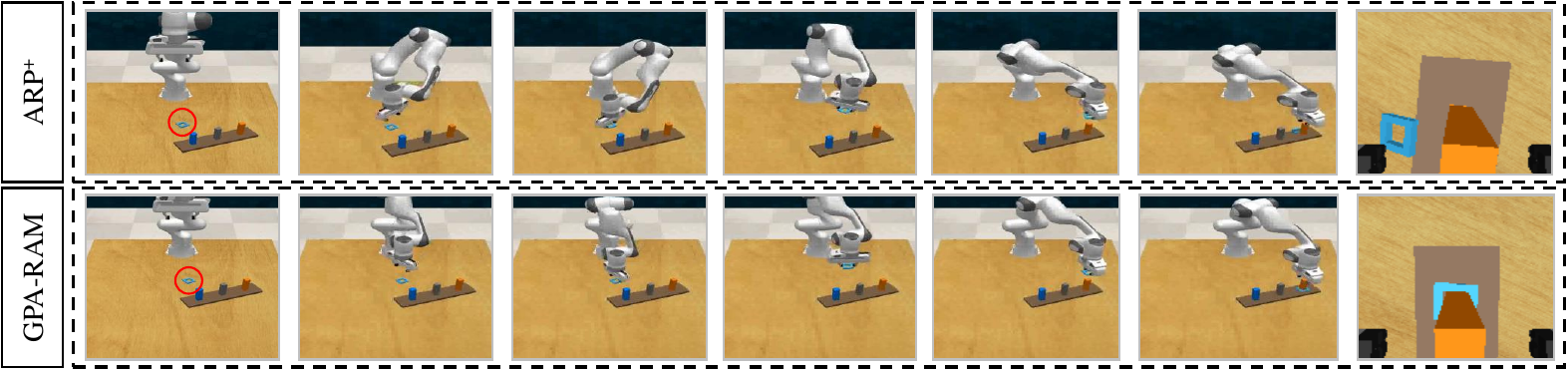}
    \caption{Qualitative Comparison of the previous SOTA (ARP$^+$) and our GPA-RAM on RLBench Tasks.
    In the top row, ARP$^+$ fails in the ``Insert the square ring onto the orange peg'' task due to an inaccurate initial grasp pose and insufficient pose adjustment.
    By contrast, GPA-RAM succeeds in this task.}
    \label{fig:rlbench_q}
\end{figure*}

\subsection{Results of RLBench-18 Tasks}

Table \ref{table:rlbench} summarizes the evaluation results for 18 RLBench tasks. In this multi-task imitation learning, the proposed GPA-RAM achieves SOTA performance in terms of both average success rate and task consistency. For the average success rate, our GPA-RAM (\textbf{87.5\%}) surpasses the previous SOTA method, ARP$^+$~\cite{arp} (\textbf{84.9\%}).
The improvement is even more pronounced when compared to the widely used RVT2~\cite{goyal2024rvt2} baseline, where our method shows an \textbf{8.2\%} increase (from \textbf{79.3\%}). This SOTA performance is reinforced by the Average Rank metric (lower is better), where our GPA-RAM (\textbf{2.28}) also achieves the top score, demonstrating superior consistency compared to ARP$^+$ (\textbf{3.28}) and RVT2 (\textbf{4.53}). Notably, the standalone RAM achieves an \textbf{84.7\%} success rate and an Average Rank of \textbf{3.22}. This performance is comparable to the previous SOTA ARP$^+$, and superior in terms of ranking, while outperforming the RVT2 baseline by \textbf{5.4\%}, thereby validating the effective spatial feature extraction capability of RAM.

This advantage in precision is also evident when comparing GPA-RAM (95.0\%) directly against ARP$^+$ (78.4\%) in the ``Insert Peg'' task. The ``Grasp-Pretraining Augmentation'' (GPA) component provides a clear benefit in other tasks demanding high-precision grasping. For example, in ``Stack Cups'', our method achieves 84.8\% success compared to ARP$^+$'s 80.0\%. Furthermore, in the challenging ``Sort Shape'' task, GPA-RAM outperforms ARP$^+$ (66.0\% vs 46.4\%), demonstrating a consistent edge in fine-grained manipulation.

The contribution of GPA was further isolated by comparing GPA-RAM with its non-GPA counterpart, RAM. Integrating GPA increased the success rates on ``Stack Cups'' and ``Sort Shape'' by \textbf{22.8 percentage points} (from 62.0\% to 84.8\%) and \textbf{11.0 percentage points} (from 55.0\% to 66.0\%), respectively. These gains show that grasp-pose priors contribute most strongly to tasks requiring precise object acquisition and alignment.

Task-level comparisons also reveal complementary behavior between grasp-focused and spatial representations. RAM achieved higher success rates on ``Place Cups'' (47.0\% versus 40.0\%), ``Push Buttons'' (96.0\% versus 89.8\%), ``Put in Cupboard'' (79.0\% versus 70.0\%), and ``Stack Blocks'' (70.4\% versus 65.6\%). These task-dependent differences are consistent with a trade-off between grasp-focused representations and the flexibility required for downstream placement, pushing, and multi-step execution. This observation motivates adaptive integration of grasp and spatial features according to downstream action requirements.

\begin{table}[t]
\caption{Evaluation on ALOHA tasks. We report success rates for Cube Transfer and Bimanual Insertion.}
\vspace{-2mm}
\label{table:aloha}
\begin{center}
    \begin{tabular}{ccccc}
    \toprule
    \multirow{2}{*}{\textbf{Model}} & \textbf{Inference Speed} & \multicolumn{3}{c}{\textbf{Cube Transfer}} \\
    \cmidrule(lr){3-5}
     & \textbf{(fps)} & \textbf{Touch} & \textbf{Lift} & \textbf{Transfer} \\
    \midrule
    BC-ConvMLP \cite{lang1} & - & 34 & 17 & 1 \\
    RT-1 \cite{RT1} & 28.7 & 44 & 33 & 2 \\
    VINN \cite{VINN} & - & 13 & 9 & 3 \\
    BeT \cite{bet} & - & 60 & 51 & 27 \\
    ACT \cite{aloha} & 61.7 & 98 & 86 & 86 \\
    RAM (ours) & \textbf{75.2} & \textbf{100} & \textbf{98} & \textbf{98} \\
    GPA-RAM (ours) & \textbf{71.3} & \textbf{100} & \textbf{100} & \textbf{98} \\
    
    \midrule
    \multirow{2}{*}{\textbf{Model}} & \textbf{Inference Speed} & \multicolumn{3}{c}{\textbf{Bimanual Insertion}} \\
    \cmidrule(lr){3-5}
    & \textbf{(fps)} & \textbf{Grasp} & \textbf{Contact} & \textbf{Insert} \\
    \midrule
    BC-ConvMLP \cite{lang1} & - & 5 & 1 & 1 \\
    RT-1 \cite{RT1} & 33.3 & 2 & 0 & 1 \\
    VINN \cite{VINN} & - & 6 & 1 & 1 \\
    BeT \cite{bet} & - & 21 & 4 & 3 \\
    ACT \cite{aloha} & 61.0 & \textbf{96} & 92 & 16 \\
    RAM (ours) & \textbf{73.0} & \textbf{96} & \textbf{94} & 32 \\
    GPA-RAM (ours) & \textbf{70.9} & \textbf{96} & \textbf{94} & \textbf{56} \\
    \bottomrule
    \end{tabular}
\end{center}
\vspace{-2mm}
\end{table}

\subsection{Results of ALOHA Bimanual Tasks} 

In Table \ref{table:aloha}, a comparison with the most competitive method (ACT \cite{aloha}) reveals that the proposed GPA-RAM delivered SOTA performance on two ALOHA bimanual tasks in terms of both success rate and inference speed. Each task is divided into three subtasks in line with the order of manipulation, and the success rate of the last subtask is equivalent to the overall task success rate. The overall success rate of the cube transferring task increased by 12\%, while the bimanual insertion task saw a remarkable improvement of 40\%. These results demonstrate the effectiveness of planning bimanual cooperation and mastering fine-grained manipulation skills. 

In addition to success rates, we explicitly evaluated the inference efficiency. As shown in Table \ref{table:aloha}, our proposed GPA-RAM achieves an inference speed of approximately 71 FPS, which surpasses the strong baseline ACT ($\sim$61 FPS) and achieves higher throughput than RT-1 ($\sim$30 FPS). This high throughput is particularly advantageous for processing continuous visual streams in dynamic environments. Unlike pure Transformer architectures, where latency spikes with increased spatial token density, RAM maintains superior efficiency, ensuring the system's responsiveness to rapid visual changes. This indicates that incorporating the grasp pretraining module enhances robustness without compromising real-time control capabilities. 

\begin{figure}[t]
    \vspace{-4mm}
    \centering
    \includegraphics[width=1.0\linewidth]{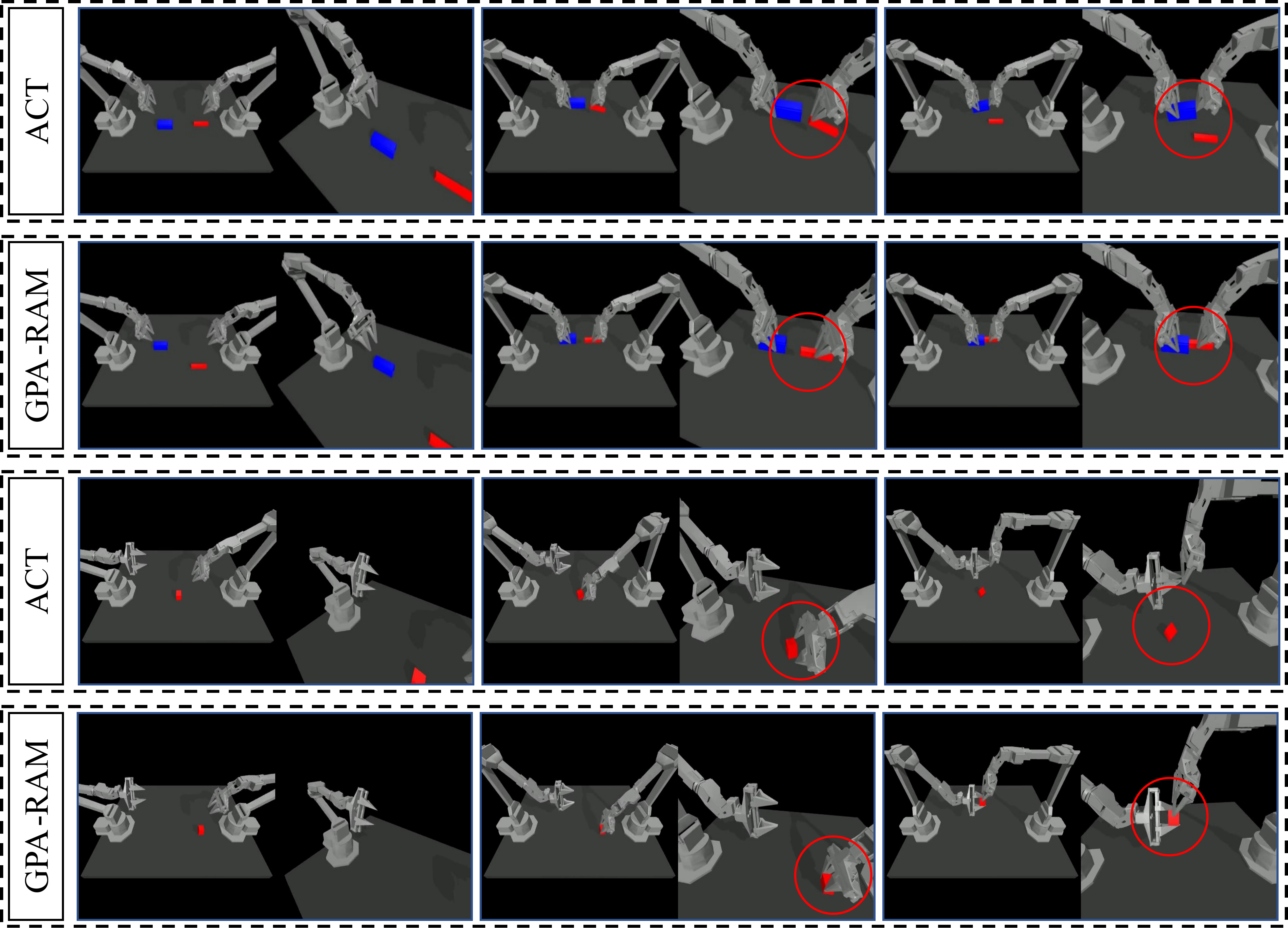}
    \caption{Qualitative Comparison of the Proposed GPA-RAM and the Previous SOTA method  (ACT) on ALOHA Bimanual Tasks. The top two rows illustrate the bimanual insertion task, where ACT struggles due to suboptimal grasping and unachievable in-hand pose adjustment of 2-finger grippers, leading to insertion failures. The bottom two rows depict the cube transfer task, where ACT fails to grasp the target cube. In contrast, GPA-RAM completes both tasks successfully, demonstrating evident advantages in execution.}
    \label{fig:aloha_q}
    \vspace{-2mm}
\end{figure}

Meanwhile, the proposed RAM and GPA-RAM excel in other subtasks, including touching, lifting, grasping, and making contact. Fig.~\ref{fig:aloha_q} provides a more qualitative comparison of GPA-RAM and the previous SOTA (ACT). Furthermore, M2T2 \cite{yuan2023m2t2} and ResNet \cite{he2016deep} were employed as grasp pose detectors for RLBench and ALOHA datasets, respectively. Since both GPAs apply in RLBench and ALOHA delivered notable boosts for manipulation, this demonstrates that our proposed GPA framework is generic and robust to different architectures of pretrained grasp pose detectors.

\subsection{Real-world Experiments}

\begin{figure}[t]
    \centering
    \includegraphics[width=1\linewidth]{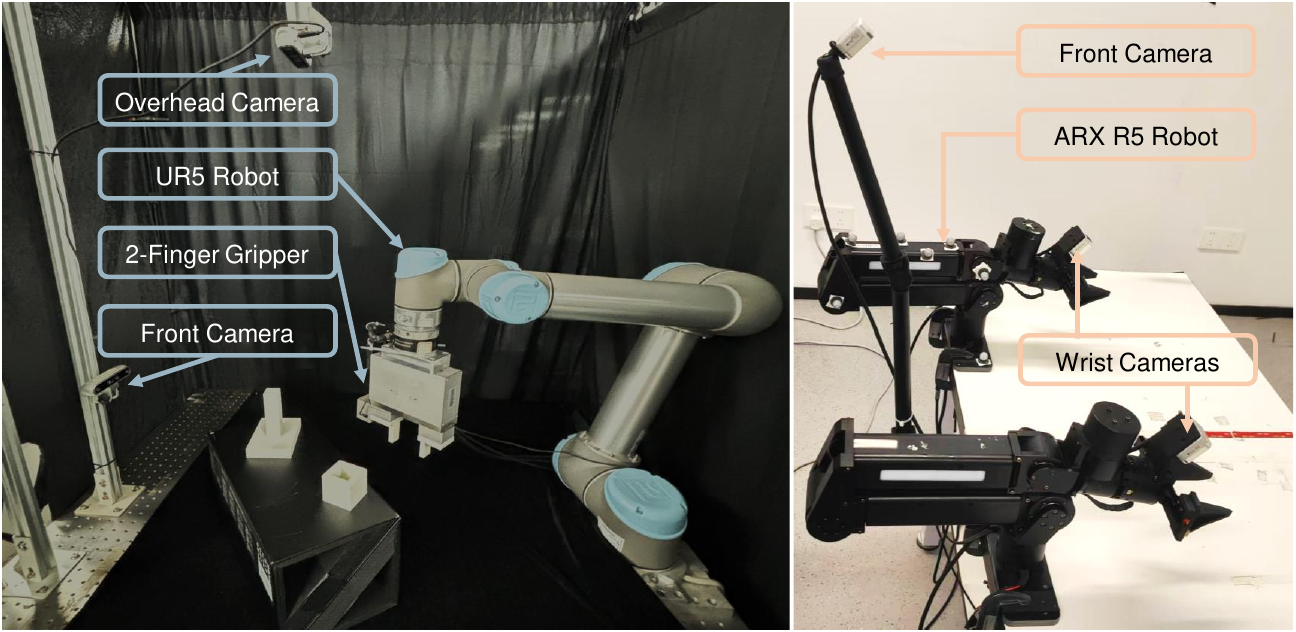}
    \caption{Real-world experimental setups for discrete action generation (left) and continuous action generation (right).}
    \label{real_setup}
    \vspace{-2mm}
\end{figure}

\textbf{Experimental Setup.}
In real-world experiments, the hardware setup was demonstrated in Fig. \ref{real_setup}. For discrete action generation, we utilized a 6-DOF UR5 manipulator with a Robust Motion GB-17-60 gripper. The vision system, adapted from the RLBench four-camera layout, employed two RealSense D455 cameras (frontal and overhead) as shown in Fig. \ref{real_setup}, with data collected via a custom joystick interface. For continuous action generation, a dual-arm ARX R5 system was employed. This setup featured three RealSense D405 cameras (mounted on the head and both wrists), with data acquisition facilitated through direct teleoperation.

\begin{figure}[ht]
    \centering
    \includegraphics[width=1\linewidth]{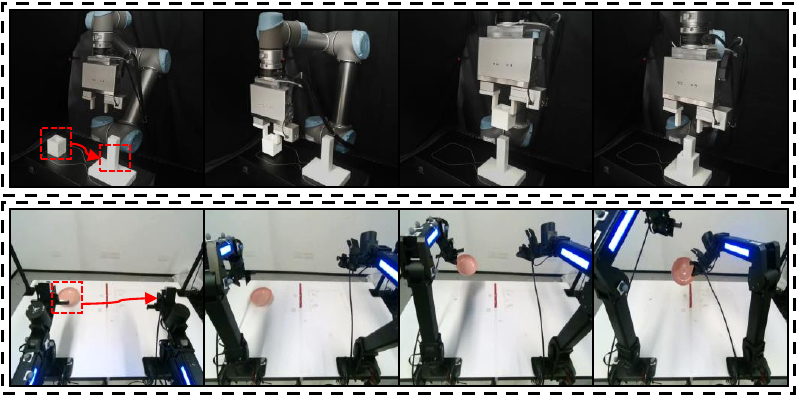}
    \caption{Examples of GPA-RAM in the real world. Top row: Grasp and insert the square hole into the square peg. Bottom row: Transfer the plate.}
    \label{fig:realword}
\end{figure}

\textbf{Real-world Tasks.} As shown in Fig. \ref{fig:realword}, Table \ref{table:real_results} and \ref{continue}, real-robot experiments involved seven discrete tasks—including block-to-drawer placement (cylindrical, hexagonal, quadrangular), bottle-to-shelf arrangement (middle and top shelves), and tight-tolerance insertion (round and square peg-in-hole)—where object poses varied significantly between training and testing. Additionally, three continuous tasks (plate transferring, tabletop organization, and bowl stacking) were deployed on the ARX R5 dual-arm robot. For more examples, see the supplementary materials.

The training dataset comprised 20 demonstrations for each discrete task and 50 demonstrations for each continuous task. Each task was tested in 10 unseen scenarios. By leveraging learning from demonstrations, our method aims to generate successful trajectories capable of handling high-precision manipulation constraints.

\begin{table}[t]
\caption{Discrete Action Tasks Performance on Real-world.}
\label{table:real_results}
\centering
\begin{tabular}{cccc}
\toprule
\textbf{Task} & \textbf{RVT2} & \textbf{ARP$^+$} &\textbf{GPA-RAM} \\
\toprule
Place cylindrical block in drawer   & 70\% & 90\% & \textbf{100\%} \\
Place quadrangular block in drawer  & 70\% & 80\% & \textbf{90\%} \\
Place hexagonal block in drawer     & 80\% & 90\% & \textbf{100\%} \\
Put bottle on middle shelf          & 60\% & 80\% & \textbf{90\%} \\
Put bottle on top shelf             & 60\% & 80\% & \textbf{80\%} \\
Grasp and insert round hole in peg  & 50\% & 50\% & \textbf{70\%} \\
Grasp and insert square hole in peg & 30\% & 40\% & \textbf{50\%} \\
\toprule
\end{tabular}
\vspace{-4mm}
\end{table}
\begin{table}[t]
\caption{Continuous Action Tasks Performance on Real-world.}
\label{continue}
\centering
\begin{tabular}{cccc}
\toprule
\textbf{Task} & \textbf{ACT} & \textbf{RAM} &\textbf{GPA-RAM} \\
\toprule
Plate transferring & 80\% & 90\% & \textbf{100\%}\\
Tabletop Organization & 70\% & 80\% & \textbf{90\%} \\
Bowl Stacking & 60\% & 80\% & \textbf{90\%} \\
\toprule
\end{tabular}
\vspace{-4mm}
\end{table}

\textbf{Real-world Experimental Results.}
The real-world experimental results in Table \ref{table:real_results} show higher observed success rates of GPA-RAM in efficiently learning diverse manipulation tasks with few demonstrations. Our method consistently outperforms both the baseline RVT2 and the competitor ARP$^+$.

In discrete tasks, GPA-RAM excelled in placing blocks of varying shapes, achieving near-perfect success rates (90\%-100\%) and demonstrating remarkable grasping adaptability. For spatial tasks like placing bottles on shelves, the method showcased its ability to plan successful paths in constrained environments. Although peg-in-hole tasks posed challenges due to stringent alignment requirements, GPA-RAM still achieved respectable success rates (70\% and 50\%), surpassing baselines by effectively handling contact-rich insertions.

In the introduced continuous action tasks, GPA-RAM exhibited exceptional temporal consistency and bimanual coordination. It achieved a 100\% success rate in plate transferring, outperforming ACT (80\%) and RAM (90\%). In complex, long-horizon tasks such as tabletop organization and bowl stacking, our method maintained a high success rate of 90\%, indicating in handling dynamic object interactions where baselines often failed to maintain stability.

\section{Ablation Study}
We conducted ablation studies of our proposed GPA-RAM in 18 RLBench tasks and 2 ALOHA bimanual cooperation tasks to investigate the effect of its key design choices.

\subsection{Ablations on RLBench-18 Tasks}
GPA-RAM set a new SOTA with an average success rate of 87.5\% across the 18 tasks in RLBench and the ablation results are presented in Table \ref{ablations1}. All ablations are trained in multi-task settings, consistent with GPA-RAM.

\textbf{RAM.} This variant excludes the GPA module from GPA-RAM, achieving an average success rate of 84.7\% across the 18 tasks in RLBench. While this exceeds RVT2 \cite{goyal2024rvt2} by 5.4\%, the method without GPA falls short by 2.8\% compared to the proposed GPA-RAM.

\textbf{C-RAM.} This configuration removes the fine stage (F-RAM) from the RAM and deploys only the single-stage RAM, achieving a success rate of 66.6\% on RLBench. 

\textbf{GPA-C-RAM.} The GPA module is integrated into the C-RAM module. Compared to C-RAM, this improved version achieves a 2.7\% improvement on the average success rate.

\textbf{GPA-RVT2.} By integrating RVT2 with the proposed general-purpose GPA framework, the task success rate improves by 4.9\% over original RVT2. Nevertheless, GPA-RVT2 still underperforms our proposed GPA-RAM model by 3.3\%.

\begin{table}[t]
\caption{Ablations of GPA-RAM on RLBench-18 Tasks.}
\label{ablations1}
\centering
\begin{tabular}{ccc}
\toprule
\textbf{Model} & \textbf{Avg. Success} & \textbf{Difference}\\
\toprule
GPA-RAM & 87.5\% & -- \\
RAM & 84.7\% & -2.8\% \\
C-RAM & 66.6\% & -20.9\% \\
GPA-C-RAM & 69.3\% & -18.2\% \\
GPA-RVT2 & 84.2\% & -3.3\% \\
\toprule
\end{tabular}

\vspace{-4mm}
\end{table}

\begin{table}[t]
\caption{Ablations of GPA-RAM on ALOHA Tasks.} 
\label{ablations2}
\centering
\begin{tabular}{c@{\hskip 3mm} c@{\hskip 3mm} c@{\hskip 3mm} c@{\hskip 3mm} c@{\hskip 3mm} c@{\hskip 3mm} }
\toprule
\multirow{2}{*}{\textbf{Frequency}} & \multirow{2}{*}{\textbf{Model}} & \multicolumn{2}{c}{\textbf{Avg. Success}} & \multicolumn{2}{c}{\textbf{Difference}} \\
\cmidrule(lr){3-4} \cmidrule(lr){5-6}
& & \textbf{Transfer} & \textbf{Insertion}  & \textbf{Transfer} & \textbf{Insertion}  \\
\midrule
\multirow{3}{*}{\textbf{50Hz}} 
& GPA-RAM     & 98\% & 56\% & --  & --   \\
& RAM        & 98\% & 32\% & -0\% & -24\% \\
& GPA-ACT        & 98\% & 38\% & -0\% & -18\% \\
\toprule
\multirow{3}{*}{\textbf{100Hz}} 
& GPA-RAM     & 90\% & 40\% & --  & --   \\
& RAM       & 88\% & 30\% & -2\% & -10\% \\
& GPA-ACT        & 86\% & 30\% & -4\% & -10\% \\
\toprule
\end{tabular}
\vspace{-4mm}
\end{table}

\subsection{Ablations on ALOHA Bimanual Tasks}
The proposed GPA-RAM is adapted into the ACT method, and achieving the best performance with success rates of 98\% and 56\% on the cube transferring and bimanual insertion tasks, respectively, at a control frequency of 50 Hz. Additionally, we conduct experiments at 100 Hz to further evaluate the model's performance under high-frequency servoing. The ablations were trained in single-task settings, summarized in Table \ref{ablations2}.

\textbf{RAM.} By removing the GPA module, the success rate for the cube transferring task stays the same as GPA-RAM, but it declines by 24\% for the bimanual insertion task. Besides, it still outperforms ACT by 12\% and 16\% on the transferring and insertion tasks, respectively.


\textbf{GPA-ACT.}
This model integrates ACT with the proposed GPA framework, achieving 12\% and 22\% higher success rates than the original ACT on the two ALOHA tasks, respectively. However, it exhibits an 18\% performance drop in the Insertion task relative to the proposed GPA-RAM. 

The cube transferring task is relatively easy such that all the ablations are competent for this task. Moreover, the proposed GPA-RAM likewise demonstrates its advantages under higher-frequency task settings (100 Hz). Despite a slight drop in success rates, it still outperforms other methods.

To conclude, the adoption of the proposed RAM and GPA can improve the success rate of diverse robotic manipulation tasks, including high-precision discrete manipulation and bimanual continuous coordination. The ablation results demonstrated that both the learning framework GPA and the architecture design RAM can enhance the performance of different network architectures and task settings.



\section{Conclusions} 
\label{sec:conclusion}

This paper presented GPA-RAM, combining grasp-prior augmentation and efficient spatial sequence modeling for fine-grained manipulation. GPA integrates grasp-pose priors from expert demonstrations into policies without extra data or pose annotation. RAM merges attention-based interaction with state-space modeling to efficiently capture local and long-range spatial dependencies. This unified framework supports discrete keyframe prediction and continuous action generation.

GPA-RAM achieved 87.5\% average success on RLBench, alongside 98\% and 56\% on ALOHA cube-transfer and bimanual-insertion tasks, operating at ~71 FPS. Physical UR5 and ARX R5 experiments demonstrated cross-configuration adaptability. Ablations confirmed the complementary benefits of GPA and RAM, showing GPA enhances both RVT2 and ACT. Future work will explore history-aware memory mechanisms for long-horizon stability, and adaptive strategies balancing grasp priors with placement and pushing flexibility.



\bibliographystyle{IEEEtran}
\bibliography{reference}

@ARTICLE{james2020rlbench,
  author={James, Stephen and Ma, Zicong and Arrojo, David Rovick and Davison, Andrew J.},
  journal={IEEE Robotics and Automation Letters}, 
  title={RLBench: The Robot Learning Benchmark \& Learning Environment}, 
  year={2020},
  volume={5},
  number={2},
  pages={3019-3026},}

@inproceedings{NIPS2017transformer,
 author = {Vaswani, Ashish and Shazeer, Noam and Parmar, Niki and Uszkoreit, Jakob and Jones, Llion and Gomez, Aidan N and Kaiser, Lukasz and Polosukhin, Illia},
 title = {Attention is All you Need},
 booktitle = {Advances in Neural Information Processing Systems},
 year = {2017},
 volume = {30},
 pages = {6000–6010},
}

@inproceedings{gu2023mamba,
title={Mamba: Linear-Time Sequence Modeling with Selective State Spaces},
author={Albert Gu and Tri Dao},
booktitle={Conference on Language Modeling},
year={2024}
}

@INPROCEEDINGS{aloha,
      title={Learning Fine-Grained Bimanual Manipulation with Low-Cost Hardware}, 
      author={Tony Z. Zhao and Vikash Kumar and Sergey Levine and Chelsea Finn},
      year={2023},
      booktitle= {Proceedings of Robotics: Science and Systems (RSS)},    
}

@INPROCEEDINGS{umi,
      title={Universal Manipulation Interface: In-The-Wild Robot Teaching Without In-The-Wild Robots}, 
      author={Cheng Chi and Zhenjia Xu and Chuer Pan and Eric Cousineau and Benjamin Burchfiel and Siyuan Feng and Russ Tedrake and Shuran Song},
      year={2024},
      booktitle= {Proceedings of Robotics: Science and Systems (RSS)}, 
}

@article{yang2023watch_and_act,
  title={Watch and Act: Learning Robotic Manipulation From Visual Demonstration},
  author={Yang, Shuo and Zhang, Wei and Song, Ran and Cheng, Jiyu and Wang, Hesheng and Li, Yibin},
  journal={IEEE Transactions on Systems, Man, and Cybernetics: Systems},
  volume={53},
  number={7},
  pages={4404-4416},
  year={2023},
  publisher={IEEE}
}

@article{zhao2023twostageinsert,
   author={Zhao, Jingdong and Wang, Zhaomin and Zhao, Liangliang and Liu, Hong},
  journal={IEEE Transactions on Industrial Electronics}, 
  title={A Learning-Based Two-Stage Method for Submillimeter Insertion Tasks With Only Visual Inputs}, 
  year={2024},
  volume={71},
  number={7},
  pages={7381-7390},
  doi={10.1109/TIE.2023.3299051}}

@inproceedings{chi2023diffusionpolicy,
	title={Diffusion Policy: Visuomotor Policy Learning via Action Diffusion},
	author={Chi, Cheng and Feng, Siyuan and Du, Yilun and Xu, Zhenjia and Cousineau, Eric and Burchfiel, Benjamin and Song, Shuran},
	booktitle={Proceedings of Robotics: Science and Systems},
	year={2023}
}

@INPROCEEDINGS{img2img1,
  author={Bharadhwaj, Homanga and Gupta, Abhinav and Kumar, Vikash and Tulsiani, Shubham},
  booktitle={2024 IEEE International Conference on Robotics and Automation (ICRA)}, 
  title={Towards Generalizable Zero-Shot Manipulation via Translating Human Interaction Plans}, 
  year={2024},
  volume={},
  number={},
  pages={6904-6911},}

@inproceedings{img2img2,
        title={Learning Manipulation by Predicting Interaction},
        author={Jia, Zeng and Qingwen, Bu and Bangjun, Wang and Wenke, Xia and Li, Chen and Hao, Dong and Haoming, Song and Dong, Wang and Di, Hu and Ping, Luo and Heming, Cui and Bin, Zhao and Xuelong, Li and Yu, Qiao and Hongyang, Li},
        booktitle= {Proceedings of Robotics: Science and Systems (RSS)},
        year={2024}
      }

@inproceedings{
video1,
title={Zero-Shot Robotic Manipulation with Pre-Trained Image-Editing Diffusion Models},
author={Kevin Black and Mitsuhiko Nakamoto and Pranav Atreya and Homer Rich Walke and Chelsea Finn and Aviral Kumar and Sergey Levine},
booktitle={The International Conference on Learning Representations ({ICLR})},
year={2024},
}

@inproceedings{
video2,
title={Unleashing Large-Scale Video Generative Pre-training for Visual Robot Manipulation},
author={Hongtao Wu and Ya Jing and Chilam Cheang and Guangzeng Chen and Jiafeng Xu and Xinghang Li and Minghuan Liu and Hang Li and Tao Kong},
booktitle={The International Conference on Learning Representations ({ICLR})},
year={2024},
}

@article{video3,
  title={GR-MG: Leveraging Partially-Annotated Data Via Multi-Modal Goal-Conditioned Policy},
  author={Li, Peiyan and Wu, Hongtao and Huang, Yan and Cheang, Chilam and Wang, Liang and Kong, Tao},
  journal={IEEE Robotics and Automation Letters},
  year={2025},
  publisher={IEEE}
}

@article{video4,
  title={Any-point trajectory modeling for policy learning},
  author={Wen, Chuan and Lin, Xingyu and So, John and Chen, Kai and Dou, Qi and Gao, Yang and Abbeel, Pieter},
  journal={arXiv preprint arXiv:2401.00025},
  year={2023}
}

@InProceedings{lang1,
  title = 	 {BC-Z: Zero-Shot Task Generalization with Robotic Imitation Learning},
  author =       {Jang, Eric and Irpan, Alex and Khansari, Mohi and Kappler, Daniel and Ebert, Frederik and Lynch, Corey and Levine, Sergey and Finn, Chelsea},
  booktitle = 	 {Proceedings of the 5th Conference on Robot Learning},
  pages = 	 {991--1002},
  year = 	 {2022},
  editor = 	 {Faust, Aleksandra and Hsu, David and Neumann, Gerhard},
  volume = 	 {164},
  series = 	 {Proceedings of Machine Learning Research},
  month = 	 {08--11 Nov},
  publisher =    {PMLR},
}

@ARTICLE{lang2,
  author={Mees, Oier and Hermann, Lukas and Burgard, Wolfram},
  journal={IEEE Robotics and Automation Letters}, 
  title={What Matters in Language Conditioned Robotic Imitation Learning Over Unstructured Data}, 
  year={2022},
  volume={7},
  number={4},
  pages={11205-11212},
}

@inproceedings{
lang3,
title={Contrastive Imitation Learning for Language-guided Multi-Task Robotic Manipulation},
author={Teli Ma and Jiaming Zhou and Zifan Wang and Ronghe Qiu and Junwei Liang},
booktitle={8th Annual Conference on Robot Learning (CoRL)},
year={2024},
}

@InProceedings{shridhar2023perceiver,
  title = 	 {Perceiver-Actor: A Multi-Task Transformer for Robotic Manipulation},
  author =       {Shridhar, Mohit and Manuelli, Lucas and Fox, Dieter},
  booktitle = {Proceedings of the Conference on Robot Learning (CoRL)},
  pages = 	 {785--799},
  volume = 	 {205},
  year = 	 {2022}
}

@inproceedings{goyal2023rvt,
  title={RVT: Robotic View Transformer for 3{D} Object Manipulation},
  author={Goyal, Ankit and Xu, Jie and Guo, Yijie and Blukis, Valts and Chao, Yu-Wei and Fox, Dieter},
  booktitle={Proceedings of the Conference on Robot Learning (CoRL)},
  year={2023},
  organization={PMLR}
}

@inproceedings{goyal2024rvt2,
  title={RVT2: Learning Precise Manipulation from Few Demonstrations},
  author={Goyal, Ankit and Blukis, Valts and Xu, Jie and Guo, Yijie and Chao, Yu-Wei and Fox, Dieter},
  booktitle={Robotics: Science and Systems},
  year={2024},
}

@article{diffusion_model,
  title={Denoising diffusion probabilistic models},
  author={Ho, Jonathan and Jain, Ajay and Abbeel, Pieter},
  journal={Advances in neural information processing systems},
  volume={33},
  pages={6840--6851},
  year={2020}
}

@inproceedings{RT1,
    title={RT-1: Robotics Transformer for Real-World Control at Scale},
    author={Anthony	Brohan and  Noah Brown and  Justice Carbajal and et al.},
    booktitle={Proceedings of Robotics: Science and Systems (RSS)},
    year={2022}
}

@article{RT2,
      title={RT-2: Vision-Language-Action Models Transfer Web Knowledge to Robotic Control}, 
      author={Anthony Brohan and Noah Brown and Justice Carbajal and et al.},
      year={2023},
      journal={arXiv preprint arXiv:2307.15818},
}

@inproceedings{
transformer_shortcoming,
title={Efficient Self-supervised Vision Transformers for Representation Learning},
author={Chunyuan Li and Jianwei Yang and Pengchuan Zhang and Mei Gao and Bin Xiao and Xiyang Dai and Lu Yuan and Jianfeng Gao},
booktitle={The International Conference on Learning Representations ({ICLR})},
year={2022},
}

@inproceedings{yuan2023m2t2,
    title={M2T2: Multi-Task Masked Transformer for Object-centric Pick and Place},
    author={Yuan, Wentao and Murali, Adithyavairavan and Mousavian, Arsalan and Fox, Dieter},
    booktitle={Conference on Robot Learning},
    volume = {229},
    pages = {3619--3630},
    year={2023}
}

@inproceedings{mujoco,
  title={MuJoCo: A physics engine for model-based control},
  author={Todorov, Emanuel and Erez, Tom and Tassa, Yuval},
  booktitle={2012 IEEE/RSJ International Conference on Intelligent Robots and Systems},
  pages={5026--5033},
  year={2012},
  organization={IEEE},
}

@inproceedings{james2022coarse,
  title={Coarse-to-fine q-attention: Efficient learning for visual robotic manipulation via discretisation},
  author={James, Stephen and Wada, Kentaro and Laidlow, Tristan and Davison, Andrew J},
  booktitle={Proceedings of the IEEE/CVF Conference on Computer Vision and Pattern Recognition (CVPR)},
  pages={13739--13748},
  year={2022}
}

@inproceedings{guhur2023hiveformer,
  title={Instruction-driven history-aware policies for robotic manipulations},
  author={Guhur, Pierre-Louis and Chen, Shizhe and Pinel, Ricardo Garcia and Tapaswi, Makarand and Laptev, Ivan and Schmid, Cordelia},
  booktitle={Proceedings of the Conference on Robot Learning (CoRL)},
  pages={175--187},
  year={2023},
  organization={PMLR}
}

@inproceedings{gervet2023act3d,
  title={Act3{D}: 3{D} feature field transformers for multi-task robotic manipulation},
  author={Gervet, Theophile and Xian, Zhou and Gkanatsios, Nikolaos and Fragkiadaki, Katerina},
  booktitle={7th Annual Conference on Robot Learning},
  year={2023}
}

@inproceedings{chen23polarnet,
    author    = {Shizhe Chen and Ricardo Garcia and Cordelia Schmid and Ivan Laptev},
    title     = {PolarNet: 3D Point Clouds for Language-Guided Robotic Manipulation},
    booktitle = {In 7th Annual Conference on Robot Learning (CoRL)},
    year      = {2023}
}

@inproceedings{bet,
 author = {Shafiullah, Nur Muhammad and Cui, Zichen and Altanzaya, Ariuntuya (Arty) and Pinto, Lerrel},
 booktitle = {Advances in Neural Information Processing Systems},
 editor = {S. Koyejo and S. Mohamed and A. Agarwal and D. Belgrave and K. Cho and A. Oh},
 pages = {22955--22968},
 publisher = {Curran Associates, Inc.},
 title = {Behavior Transformers: Cloning k modes with one stone},
 volume = {35},
 year = {2022}
}

@article{VINN,
  title={The surprising effectiveness of representation learning for visual imitation},
  author={Pari, Jyothish and Shafiullah, Nur Muhammad and Arunachalam, Sridhar Pandian and Pinto, Lerrel},
  journal={arXiv preprint arXiv:2112.01511},
  year={2021}
}

@inproceedings{he2016deep,
  title={Deep residual learning for image recognition},
  author={He, Kaiming and Zhang, Xiangyu and Ren, Shaoqing and Sun, Jian},
  booktitle={Proceedings of the IEEE/CVF Conference on Computer Vision and Pattern Recognition (CVPR)},
  pages={770--778},
  year={2016}
}

@inproceedings{radford2021clip,
  title={Learning transferable visual models from natural language supervision},
  author={Radford, Alec and Kim, Jong Wook and Hallacy, Chris and Ramesh, Aditya and Goh, Gabriel and Agarwal, Sandhini and Sastry, Girish and Askell, Amanda and Mishkin, Pamela and Clark, Jack and others},
  booktitle={International Conference on Machine Learning (ICML)},
  pages={8748--8763},
  year={2021}
}

@ARTICLE{grasp1,
  author={Zhong, Xungao and Gong, Tao and Yu, Junzhi and Zhou, Chengxian and Zhong, Xunyu and Liu, Qiang},
  journal={IEEE Transactions on Automation Science and Engineering}, 
  title={RIGNet: Robot Intention Grasp for Dense Stacked Targets With Multi-Task Siamese Schema Through RoIs Learning}, 
  year={2025},
  volume={22},
  number={},
  pages={10354-10367},}

@ARTICLE{grasp2,
  author={Tang, Chao and Huang, Dehao and Dong, Wenlong and Xu, Ruinian and Zhang, Hong},
  journal={IEEE Transactions on Automation Science and Engineering}, 
  title={FoundationGrasp: Generalizable Task-Oriented Grasping With Foundation Models}, 
  year={2025},
  volume={22},
  number={},
  pages={12418-12435},
  doi={10.1109/TASE.2025.3542418}}

@ARTICLE{grasp3,
  author={Xu, Peng and Cheng, Hu and Wang, Jiankun and Meng, Max Q.-H.},
  journal={IEEE Transactions on Automation Science and Engineering}, 
  title={Learning to Reorient Objects With Stable Placements Afforded by Extrinsic Supports}, 
  year={2024},
  volume={21},
  number={4},
  pages={5653-5664},
  doi={10.1109/TASE.2023.3314461}}

@ARTICLE{v1,
  author={Li, Xin and He, Bin and Wang, Zhipeng and Zhou, Yanmin and Li, Gang and Li, Xiang},
  journal={IEEE Transactions on Automation Science and Engineering}, 
  title={Toward Cognitive Digital Twin System of Human-Robot Collaboration Manipulation}, 
  year={2025},
  volume={22},
  number={},
  pages={6677-6690},
  doi={10.1109/TASE.2024.3452149}}

@ARTICLE{tro1,
  author={Xu, Kechun and Zhou, Zhongxiang and Wu, Jun and Lu, Haojian and Xiong, Rong and Wang, Yue},
  journal={IEEE Transactions on Robotics}, 
  title={Grasp, See, and Place: Efficient Unknown Object Rearrangement With Policy Structure Prior}, 
  year={2025},
  volume={41},
  number={},
  pages={464-483},
  doi={10.1109/TRO.2024.3502520}}

@ARTICLE{fp2at,
  author={Liu, Yangjun and Liu, Sheng and Chen, Binghan and Yang, Zhi-Xin and Xu, Sheng},
  journal={IEEE Transactions on Robotics}, 
  title={Fusion-Perception-to-Action Transformer: Enhancing Robotic Manipulation With 3-D Visual Fusion Attention and Proprioception}, 
  year={2025},
  volume={41},
  number={},
  pages={1553-1567},
  doi={10.1109/TRO.2025.3539193}}

@ARTICLE{arp,
  author={Zhang, Xinyu and Liu, Yuhan and Chang, Haonan and Schramm, Liam and Boularias, Abdeslam},
  journal={IEEE Robotics and Automation Letters}, 
  title={Autoregressive Action Sequence Learning for Robotic Manipulation}, 
  year={2025},
  volume={10},
  number={5},
  pages={4898-4905},
  doi={10.1109/LRA.2025.3550849}}

@misc{gu2022efficientlymodelinglongsequences,
      title={Efficiently Modeling Long Sequences with Structured State Spaces}, 
      author={Albert Gu and Karan Goel and Christopher Ré},
      year={2022},
      eprint={2111.00396},
      archivePrefix={arXiv},
      primaryClass={cs.LG},
      url={https://arxiv.org/abs/2111.00396}, 
}

@ARTICLE{tmech1,
  author={Zeng, Chao and Li, Zhan and Yang, Yipeng and Liu, Cunjia and Yang, Chenguang},
  journal={IEEE/ASME Transactions on Mechatronics}, 
  title={Learning-Based Methods for Aerial Manipulation: A Focused Review}, 
  year={2026},
  volume={31},
  number={2},
  pages={2270-2287},
  doi={10.1109/TMECH.2025.3625915}}

@ARTICLE{tmech-dp,
  author={Zhang, Hao and Kan, Zhen and Shang, Weiwei and Song, Yongduan},
  journal={IEEE/ASME Transactions on Mechatronics}, 
  title={A Novel Task-Driven Diffusion-Based Policy With Affordance Learning for Generalizable Manipulation of Articulated Objects}, 
  year={2026},
  volume={31},
  number={2},
  pages={1241-1253},
  doi={10.1109/TMECH.2025.3602121}}

@ARTICLE{tmech2,
  author={Yao, Xiangtong and Blei, Tobias and Meng, Yuan and Zhang, Yu and Zhou, Hongkuan and Bing, Zhenshan and Huang, Kai and Sun, Fuchun and Knoll, Alois},
  journal={IEEE/ASME Transactions on Mechatronics}, 
  title={Long-Horizon Language-Conditioned Imitation Learning for Robotic Manipulation}, 
  year={2025},
  volume={30},
  number={6},
  pages={5628-5639},
  doi={10.1109/TMECH.2025.3547047}}

@ARTICLE{tmech3,
  author={Zhao, Shuqi and Zhu, Xinghao and Chen, Yuxin and Li, Chenran and Xie, Yichen and Zhang, Xiang and Ding, Mingyu and Tomizuka, Masayoshi},
  journal={IEEE/ASME Transactions on Mechatronics}, 
  title={DexH2R: Task-Oriented Dexterous Manipulation From Human to Robots}, 
  year={2026},
  volume={31},
  number={3},
  pages={3202-3213},
  doi={10.1109/TMECH.2025.3641164}}

@ARTICLE{tmech4,
  author={Sun, Fuchun and Miao, Shengyi and Zhong, Daming and Wu, Lianghong and Zhou, Huaidong and Wang, Na and Wen, Zhenkun and Huang, Haiming},
  journal={IEEE/ASME Transactions on Mechatronics}, 
  title={Manipulation Skill Representation and Knowledge Reasoning for 3C Assembly}, 
  year={2025},
  volume={30},
  number={6},
  pages={5387-5397},
  doi={10.1109/TMECH.2025.3543739}}

@ARTICLE{tmech5,
  author={Liu, Jin and Sun, Kai and Miao, Yanzi and Wang, Chaoqun and Wang, Hesheng},
  journal={IEEE/ASME Transactions on Mechatronics}, 
  title={Geometry-Guided Self-Supervised Learning for Category-Level Object Pose Estimation in Robotic Grasping}, 
  year={2026},
  volume={31},
  number={3},
  pages={3179-3189},
  doi={10.1109/TMECH.2025.3642869}}

\end{document}